\documentclass{article}
\usepackage{spconf,amsmath,graphicx}

\usepackage{graphicx}
\usepackage{multicol}
\usepackage{multirow}
\usepackage{makecell}
\usepackage{amsmath}
\usepackage{algorithm}
\usepackage{algorithmic}
\usepackage{xcolor,marginnote,colortbl}
\usepackage{setspace}
\usepackage{nicefrac}
\usepackage{float}
\usepackage{url}

\usepackage{breakurl}

\title{Data-free mixed-precision quantization using novel sensitivity metric}
\name{Donghyun Lee\sthanks{Equal contribution}, Minkyoung Cho\footnotemark[1], Seungwon Lee, Joonho Song, and Changkyu Choi}
\address{Samsung Advanced Institute of Technology, Samsung Electronics, South Korea}
\begin{document}
\twocolumn[
\begin{@twocolumnfalse}
© 2021 IEEE. Personal use of this material is permitted. Permission from IEEE must be obtained for all other uses, in any current or future media, including reprinting/republishing this material for advertising or promotional purposes, creating new collective works, for resale or redistribution to servers or lists, or reuse of any copyrighted component of this work in other works.
\end{@twocolumnfalse}
]
\maketitle
\begin{abstract}
Post-training quantization is a representative technique for compressing neural networks, making them smaller and more efficient for deployment on edge devices. However, an inaccessible user dataset often makes it difficult to ensure the quality of the quantized neural network in practice. In addition, existing approaches may use a single uniform bit-width across the network, resulting in significant accuracy degradation at extremely low bit-widths. To utilize multiple bit-width, sensitivity metric plays a key role in balancing accuracy and compression. In this paper, we propose a novel sensitivity metric that considers the effect of quantization error on task loss and interaction with other layers. Moreover, we develop labeled data generation methods that are not dependent on a specific operation of the neural network. Our experiments show that the proposed metric better represents quantization sensitivity, and generated data are more feasible to be applied to mixed-precision quantization.

\end{abstract}
\section{Introduction}
\label{sec:intro}

\begin{figure}[t]
\begin{minipage}[b]{.48\linewidth}
  \centering
  \centerline{\includegraphics[width=4.0cm]{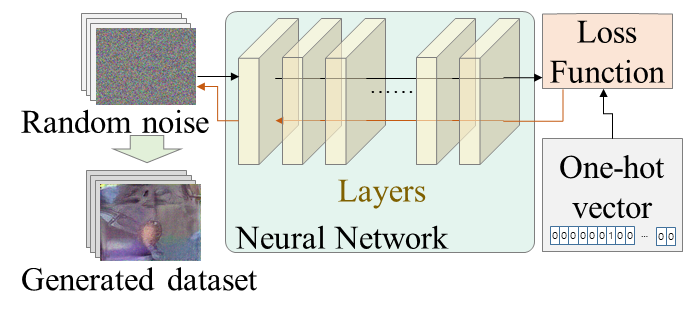}}
  \centerline{(a) Data generation}\medskip
\end{minipage}
\begin{minipage}[b]{.48\linewidth}
  \centering
  \centerline{\includegraphics[width=4.0cm]{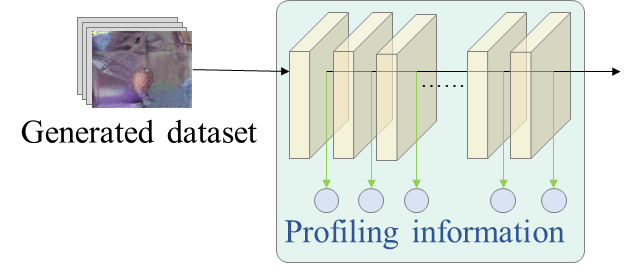}}
  \centerline{(b) Profiling for quantization}\medskip
\end{minipage}
\hfill
\begin{minipage}[b]{0.48\linewidth}
  \centering
  \centerline{\includegraphics[width=4.0cm]{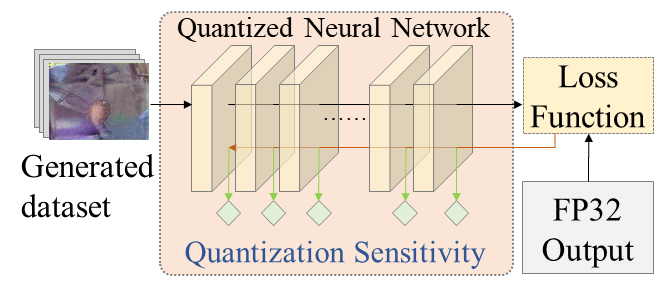}}
  \centerline{(c) Sensitivity measurement}\medskip
\end{minipage}
\begin{minipage}[b]{0.48\linewidth}
  \centering
  \centerline{\includegraphics[width=4.0cm]{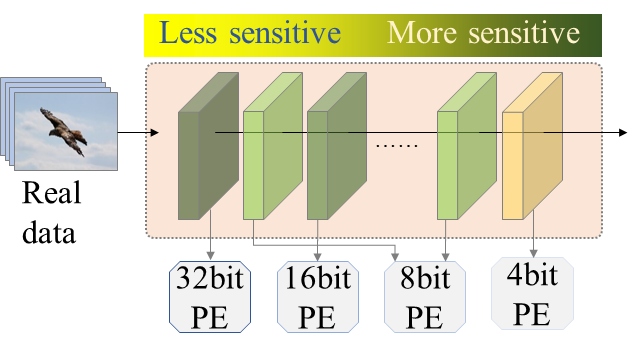}}
  \centerline{(d) Mixed-precision inference}\medskip
\end{minipage}
\caption{Overall process of a post-training mixed-precision quantization method. (a) Generate dataset from pretrained network. (b) Post-training quantization using statistics. (c) Measure the sensitivity of each layer. (d) Quantize to higher precision for a more sensitive layer.}
\label{fig:concept}
\end{figure}

In recent years, deep neural networks have simplified and enabled applications in numerous domains, especially for vision tasks \cite{szegedy2016rethinking, liu2016ssd, chen2018encoder}. Meanwhile, there is a need to minimize the memory footprint and reduce the network computation cost to deploy on edge devices. Significant efforts have been made to reduce the network size or accelerate inference of the neural network \cite{mv3, hinton2015distilling, han2015deep}. Several approaches exist to this problem, and quantization has been studied as one of the most reliable solutions. In the quantization process, low-bit representations of both weights and activations introduce quantization noise, which results in accuracy degradation. To alleviate the accuracy loss, retraining or fine-tuning methods are developed by exploiting extensive training datasets \cite{ qil2019, jchoi2018pact, dongqing2018lqnets}. 

However, these methods are not applicable in many real-world scenarios, where the user dataset is inaccessible due to confidential or personal issues \cite{zeroq2020}. It is impossible to train the network and verify the quality of a quantized neural network without a user dataset. Although post-training quantization is a frequently suggested method to address this problem \cite{nayak2019bit, mnagel2019dfq}, a small dataset is often required to decide the optimal clipping range of each layer. The credential statistics of layers are important to ensure the performance of the quantized network in increasing task loss at ultra-low-bit precision.

Considering that most quantization approaches have used uniform bit allocation across the network, the mixed-precision quantization approach takes a step further in preserving the accuracy by lifting the limit of those approaches \cite{kwang2019haq, mpq2018}. As a necessity, several sensitivity measurement methods for maximizing the effects of mixed-precision have also been proposed because it is difficult to determine which sections of the network are comparatively less susceptible to quantization \cite{zhe2019optimizing, hawqv2, molchanov2019importance}. To measure the quantization robustness of activations and weights, it is necessary to analyze the statistics generated during forward and backward processes by using a reliable dataset. The prior sensitivity metrics use the difference between outputs of the original model and the quantized model when quantizing each layer separately \cite{zeroq2020, zhe2019optimizing}. However, this approach does not consider the interaction of quantization error and other layers. It cannot be neglected because a lower bit quantization implies a larger quantization error. Other prior studies require severe approximations to compute higher-order derivatives efficiently \cite{hawqv2, zdong2019hawq}. 

 In this work, we provide a straightforward method to compute the layer-wise sensitivity for mixed-precision quantization, considering the interaction of quantization error. In addition, we propose a data generation method, which is effective in the process of post-training mixed-precision quantization, as shown in Figure~\ref{fig:concept}. Prior works \cite{zeroq2020, yin2020dreaming} use statistics of a particular operation (such as batch normalization), and it is impotent when the networks that do not have the specific operation explicitly. The proposed synthetic data engineering approach is independent of network structures. Moreover, it generates labeled data to verify the quantized network.

The remainder of this paper is organized as follows. Section \ref{sec:method} describes the data generation method and proposes a sensitivity measurement metric. Section \ref{sec:experiments} provides an experimental demonstration of mixed-precision quantization using the proposed metric and generated data, and the results are compared with previous approaches.


\section{Methodology}
\label{sec:method}

\begin{figure}[t]
\centering
  \includegraphics[width=0.9\linewidth]{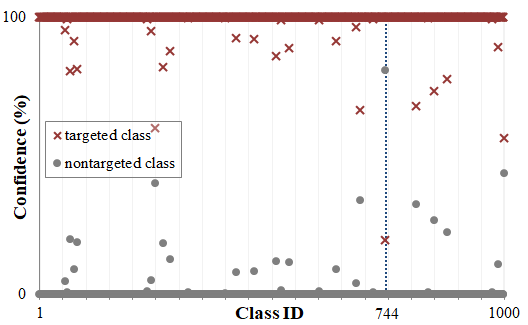}
    \caption{Top-2 confidences of each generated data sample on ResNet50. Confidence means the output values of the softmax layer. For all classes except one (Class 744), the generated labeled data samples pointed their target classes corresponding to the labels with the highest confidence.}
    \label{fig:confidence_rn50}
\end{figure}

\subsection{Data Generation}
\label{ssec:datagen}
For a general data generation method, we seek to avoid any dependency on a specific operation in a convolutional network. As shown in Figure~\ref{fig:concept}(a), we first forward a noisy image initialized from a uniform distribution in the range between 0 and 255, inclusive. To produce a set of labeled data, we use a set of class-relevant vectors/matrices, which means one-hot vectors per class in the classification task. In the forward pass, the loss is computed as

\begin{align}
    {x_{c}^{\ast}}=\underset{x_{c}}{\operatorname{argmin}}\,\mathcal{L}_{CE}(f(y({x_{c}})),\,{v_{c}})\,+\,\lambda\cdot{y}({x_{c}})
    \label{eq:datagen}
\end{align}
where ${x_{c}}$ is the trainable input feature, ${v_{c}}$ is the set of class-relevant vectors/matrices, called the golden set, $y(\cdot)$ is the neural network, and $f(\cdot)$ is an activation function (i.e., softmax) used to compute cross-entropy loss ($\mathcal{L}_{CE}$) with the golden set. In addition, we reinforce the loss by maximizing the activation of output of network to enhance the efficiency of data generation process by referring the prior works for model interpretation \cite{deepdream, visualize}. 
The calculated loss is propagated in the backward pass and generates input feature $x_{c}^{\ast}$ for each class, $c \in\{1 \ldots NumOfClass\}$. Finally, we have crafted synthetic data for each class after several iterations of forward and backward processes.

Figure~\ref{fig:confidence_rn50} demonstrates the reliability of crafting labeled data for each class by showing the top-2 confidences per class of synthetic data. All the generated data are used not only to measure the layer-wise sensitivity of the network but also to obtain the statistics of activations to improve the quality of the post-training quantized network.

\subsection{Sensitivity Metric}
\label{ssec:sensmetric}

The objective of mixed-precision quantization is to allocate appropriate bits to each layer to reduce the cost (e.g., bit operations~\cite{van2020bayesian}) of neural network models while suppressing the task loss growth. The sensitivity metric is an important factor for finding the optimal point between the effects of quantization error and cost. First, we would like to measure the effect of quantization error on the loss of a network that has total $L$ layers when weights of $i^{th}$ layer $(1 \leq i \leq L)$, $W_i$ are quantized through quantization function $Q_{k}(\cdot)$ into $k$-bit. Given input data $x$ and quantized neural network $\hat{y}$, we consider the Euclidean distance between $y(x)$ and $\hat{y}(x)$ as the $\mathcal{L}$, the loss of network output, for sensitivity measurement instead of task loss. We can define the effect of quantization error $Q_{k}(W_i) - W_i$ of $Q_{k}(W_i)$ on $\mathcal{L}$ as follows:

\begin{align}
    \Omega_{W_i}(k) = \frac1N {\sum_{x}\biggl\lVert\frac{\partial \mathcal{L}}{\partial (Q_{k}(W_i) - W_i)}\biggr\rVert}
    \label{eq:s1}
\end{align}

\noindent where $N$ denotes the total size of the dataset, and $\Omega_{W_i}(k)$ are gradients for the quantization error. $W_i$ is not variable in post-training quantization; thus, we can represent Eq.~\ref{eq:s1} as weight gradients of quantized parameters by using the chain rule as follows:

\begin{align}
    \frac{\partial \mathcal{L}}{\partial Q_{k}(W)} \cdot \frac{\partial Q_{k}(W)}{\partial (Q_{k}(W) - W)}
    \simeq \frac{\partial \mathcal{L}}{\partial Q_{k}(W)}
    \label{eq:s2}
\end{align}

The effects of the activation’s quantization error on $\mathcal{L}$ is represented as activation gradients of quantized activation by applying the derivation of the formula shown in the previous equations. Subsequently, we calculate the expectation for the effects of the quantized network on $\mathcal{L}$ by using the geometric mean of $\Omega$ of ${m}$-bit quantized activation $Q_{m}(A)$ and quantized weights $Q_{k}(W)$, which is formulated as

\begin{align}
    E[|S_{\hat{y}}|]
    = \biggl(\prod_i^L \Omega_{A_i}(m_i) \Omega_{W_i}(k_i)\biggr)^\frac1L
    \label{eq:s3}
\end{align}

The gradients of the converged single-precision neural network are not changed for the same data. To measure the effect of $Q_{k}(W_i) - W_i$ on other connected layers, we observe the gradient perturbations of $W$ and $A$, which are caused by $Q_{k}(W_i)$. Consequently, we can measure the effect of quantization error on other layers and loss of the network together, i.e., sensitivity of quantization by using $E[|S_{\hat{y}}|]$ when we only quantize activations or weights of a layer for which we would like to measure the sensitivity. Expressing the single-precision as $Q_{\mathrm{FP32}}(\cdot)$, we have

\begin{align*}
    \tag{5}
    E[|S_{W_j}|] & = \biggl(\prod_i^L \Omega_{A_i}(\mathrm{FP32}) \Omega_{W_i}(k_i) \biggr)^\frac1L \\
    s.t. \quad & k_i = 
    \begin{cases}
        k_i < 32 &  i = j\\
        \mathrm{FP32} & i \neq j
    \end{cases}
\label{eq:s4}
\end{align*}

\noindent for quantization sensitivity metric for $W_j$. It is straightforward by using the information of back-propagation of quantized network and considers the effect of quantization error on other layers.


\section{Experiments}
\label{sec:experiments}

In this section, we first demonstrate the effectiveness of the proposed data generation method in post-training quantization. Then, we show that the proposed sensitivity metric represents the quantization sensitivity of the layer effectively by using our generated data. Finally, sensitivity value by using various datasets indicates that the proposed data generation method is also credible in sensitivity measurement.

To demonstrate our methodology, we use classification models VGG19, ResNet50, and InceptionV3 on the ImageNet validation dataset.

\begin{table}[ht]
  \centering
  \begin{tabular}{clrr}
    \hline
    \rowcolor{lightgray} \textbf{Model} & \textbf{Dataset} & \textbf{Top-1} & \textbf{Top-5} \\ \hline
    \multirow{4}{*}{VGG19}
    & ImageNet & 72.31 & 90.81  \\ \cline{2-4}
    & Noise & 3.45 & 10.45  \\
    & ZeroQ & - & -  \\
    & Proposed & 71.84 & 90.53  \\ \hline
    \multirow{4}{*}{ResNet50}
    & ImageNet & 75.89 & 92.74  \\ \cline{2-4}
    & Noise & 11.28 & 29.88  \\
    & ZeroQ & 75.47 & 92.62  \\
    & Proposed & 75.68 & 92.72 \\ \hline
    \multirow{4}{*}{InceptionV3}
    & ImageNet & 77.14 & 93.43  \\ \cline{2-4}
    & Noise & 62.49 & 85.00  \\
    & ZeroQ & 76.85 & 93.31  \\
    & Proposed & 76.83 & 93.26  \\ \hline
  \end{tabular}
  \caption{Results of post-training quantization (8-bit quantization for activations and weights) using different dataset. ImageNet presents the oracle performance in terms of using training dataset and others are the results of the data-free methods.}
  \label{tbl:expe_datagen_table}
\end{table}

\subsection{Efficacy of Data Generation Method }
\label{ssec:expe_datagen}
We evaluate our method using the generated data to determine the clipping range of each layer in post-training quantization. Our experiments exploit a simple symmetric quantization approach, which uses the maximum value of activation as the clipping range of each layer. Hence, maximum values are extremely crucial for confining the dynamic range to utilize the bit-width efficiently, preventing a severe accuracy drop in low-bit quantization.

For our proposed method, input images are initialized according to uniform distribution, which follows standardization or normalization. To generate data, we use Adam optimizer to optimize the loss function with a learning rate of 0.04, and we found that $\lambda$=1 works best empirically.

For InceptionV3, input data are initialized according to $\mathcal{U}$(0, 255), which follows standardization with mean and variance by considering that the model requires synthetic data to be converted into the range of [-1, 1] through preprocessing. For VGG19 and ResNet50, input data are initialized according to $\mathcal{U}$(0, 255), which follows normalization with the factor of 255 because they require the range of [0, 1].

Table~\ref{tbl:expe_datagen_table} shows the empirical results for ImageNet, random noise, and ZeroQ \cite{zeroq2020}. In the experiment, ImageNet data are produced by choosing one image per class from the training data, and we generate 1000 data samples randomly using the existing method and the proposed method. As one can see, our method shows similar or higher performances over existing data-free method, having less than 0.5\% differences from the results of ImageNet cases. As shown in VGG19, although the existing research is hard to generalize, the method maintains sound performance regardless of the network structure.

\subsection{Comparison of Sensitivity Metrics}
\label{ssec:expe_sensitivity}

\begin{figure*}[ht]
    \centering
        \begin{minipage}[b]{0.24\linewidth}
          \centering
          \centerline{\includegraphics[width=4cm]{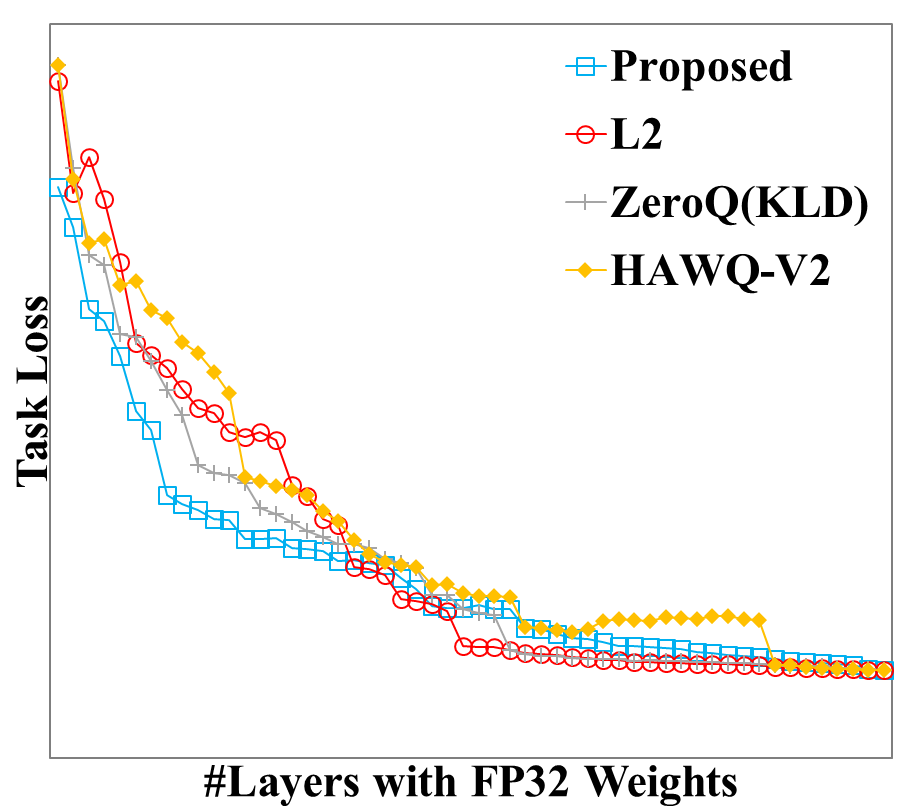}}
          \centerline{(a) ResNet50 A32W\{32,4\}}\medskip
        \end{minipage}
        \begin{minipage}[b]{0.24\linewidth}
          \centering
          \centerline{\includegraphics[width=4cm]{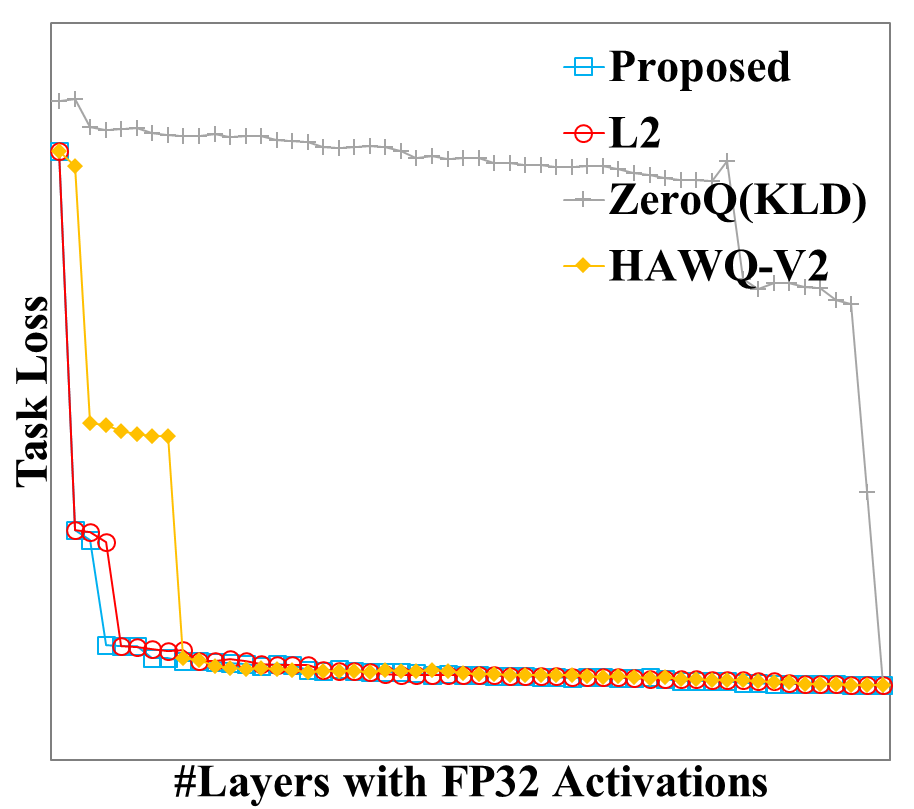}}
          \centerline{(b) ResNet50 A\{32, 4\}W32}\medskip
        \end{minipage}
        \begin{minipage}[b]{0.24\linewidth}
          \centering
          \centerline{\includegraphics[width=4cm]{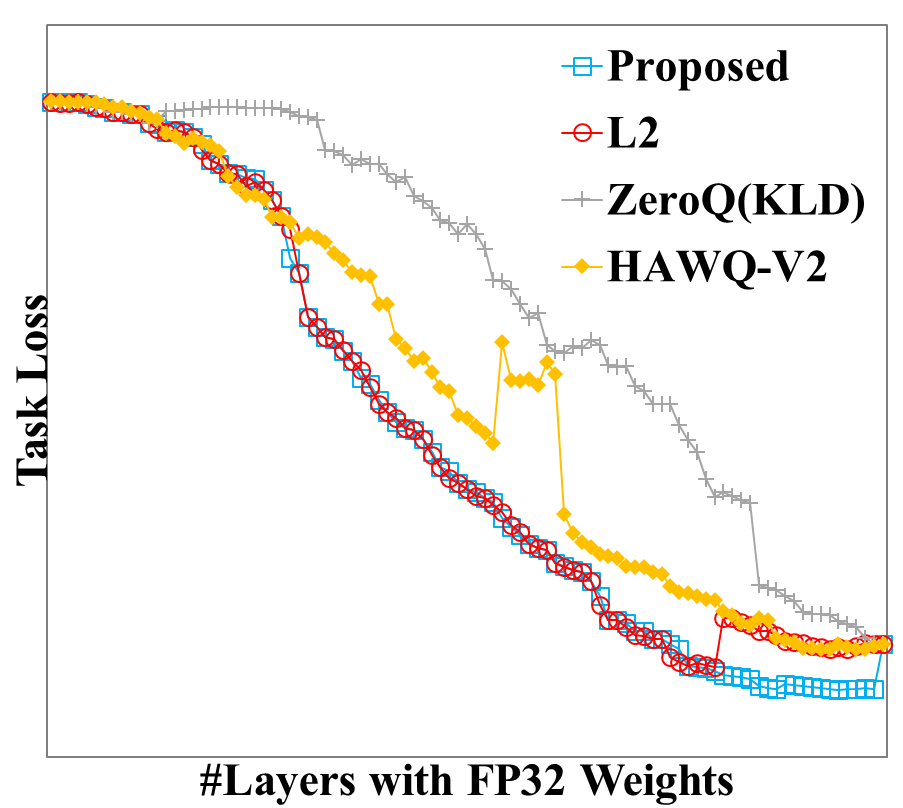}}
          \centerline{(c) InceptionV3 A32W\{32,4\}}\medskip
        \end{minipage}
        \begin{minipage}[b]{0.24\linewidth}
          \centering
          \centerline{\includegraphics[width=4cm]{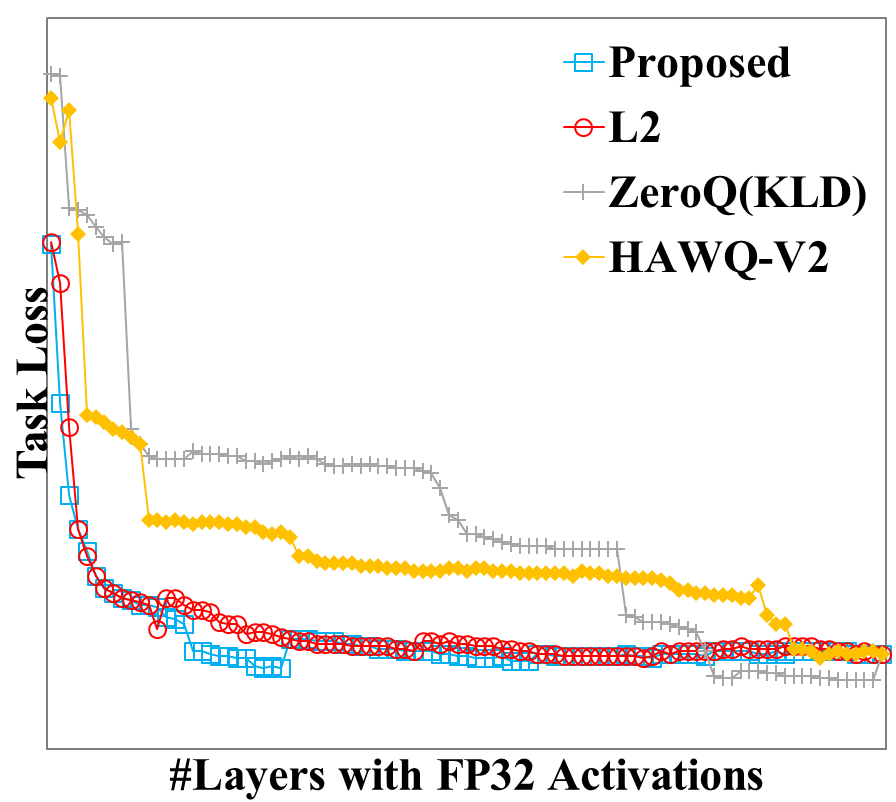}}
          \centerline{(d) InceptionV3 A\{32, 4\}W32}\medskip
        \end{minipage}
    \caption{Results of sensitivity metric evaluation on ResNet50 and InceptionV3 over ImageNet dataset. A32W\{32,4\} is the evaluation for weight sensitivity and A\{32,4\}W32 is for activation sensitivity.}
    \label{fig:expe_sensitivity}
\end{figure*}

To verify several metrics in weight sensitivity, we measure the task loss by switching floating-point weights in the order of layers with higher sensitivity values, where all weights are quantized to 4-bit, and activations do not have quantization error. We denote this experiment as A32W\{32,4\}. A\{32,4\}W32 indicates the evaluation of the metrics in activation sensitivity when switching the 4-bit quantized activation to floating-point, where weights are single-precision. ZeroQ~\cite{zeroq2020} measures the KL divergence and~\cite{zhe2019optimizing} measures the Euclidean distance between the original model and the quantized model. ZeroQ~\cite{zeroq2020} only measures the weight sensitivity. HAWQ-V2~\cite{hawqv2} uses the average Hessian trace and L2 norm of quantization perturbation. We use the data generated by the proposed method for all metrics.

Figure~\ref{fig:expe_sensitivity} shows the results of ResNet50 and InceptionV3 over ImageNet dataset. Our sensitivity metric reliably and quickly lowered the task loss, which means that the proposed sensitivity metric is good at weights and activations that are comparatively more susceptible to quantization. To express the results of Figure~\ref{fig:expe_sensitivity} quantitatively, we calculate the relative area under the curve of task loss graph considering the result of the proposed metric as 1 and summarize it in Table~\ref{tbl:expe_sensitivity_table}. Our proposed method has the smallest area in all results.

\begin{table}[t]
  \centering
  \begin{tabular}{m{0.18\linewidth}m{0.2\linewidth}rr}
    \hline
    \rowcolor{lightgray} \textbf{Model} & \textbf{Metric} & \textbf{A32W\{32,4\}} & \textbf{A\{32,4\}W32} \\ \hline
    \multirow{4}{*}{ResNet50}
    & Proposed & 1  & 1\\ \cline{2-4}
    & L2~\cite{zhe2019optimizing} & 1.22 & 1.11 \\
    & ZeroQ(KLD) & 1.14 & 18.48 \\
    & HAWQ-V2 & 1.42 & 2.09 \\ \hline
    \multirow{4}{*}{InceptionV3}
    & Proposed & 1 & 1 \\ \cline{2-4}
    & L2 & 1.04 & 1.50 \\
    & ZeroQ(KLD) & 1.69 & 8.06 \\
    & HAWQ-V2 & 1.25 & 6.17 \\ \hline
  \end{tabular}
  \caption{Quantitative comparison of different sensitivity metrics through relative area calculation of task loss graph.}
  \label{tbl:expe_sensitivity_table}
\end{table}

\subsection{Sensitivity According to Dataset Type}
\label{ssec:expe_datagen_sensitivity}

To show the importance of the data in measuring the sensitivity, we evaluate the proposed sensitivity metric over the different datasets of Section~\ref{ssec:expe_datagen} on the 4-bit quantized network that clipped the activation range using ImageNet dataset. We measure the task loss as in Section~\ref{ssec:expe_sensitivity}. The proposed dataset is the most similar in sensitivity to the ImageNet dataset, as shown in Table~\ref{tbl:expe_datagen_sensitivity_table}.
Notably, the proposed data generation method provides reliable statistics similar to the original training dataset. For InceptionV3, preprocessed ImageNet data are in the range of [-2.03, 2.52]. However, the range of the preprocessed data from~\cite{zeroq2020} is measured as [-11.50, 11.21], while that of our data is in [-2.35, 2.36], whose maximum value is almost similar to that of the ImageNet dataset. These similar statistics are seen in ResNet50. The incorrect statistics of the activation data corresponding to the original training dataset implies inaccurate sensitivity measurement~\cite{zeroq2020}. Thus, our generation method can be used for reliable sensitivity measurement. 

\begin{table}[t]
  \centering
  \begin{tabular}{clrr}
    \hline
    \rowcolor{lightgray} \textbf{Model} & \textbf{Dataset} & \textbf{A32W\{32,4\}} & \textbf{A\{32,4\}W32} \\ \hline
    \multirow{4}{*}{ResNet50}
    & Proposed & 1 & 1 \\ \cline{2-4}
    & ImageNet & 0.74 & 1.60 \\ 
    & Noise & 1.35 & 6.76 \\ 
    & ZeroQ & 1.11 & 3.17 \\ \hline 
    \multirow{4}{*}{InceptionV3}
    & Proposed & 1 & 1 \\ \cline{2-4}
    & ImageNet & 1.07 & 2.48 \\ 
    & Noise & 1.36 & 3.54 \\
    & ZeroQ & 1.47 & 2.63 \\ \hline
  \end{tabular}
  \caption{Quantitative comparison of using different datasets in sensitivity measurement through relative area calculation of task loss graph.}
  \label{tbl:expe_datagen_sensitivity_table}
\end{table}


\section{Conclusion}
\label{sec:conclusion}

In this paper, we proposed an effective data generation method for post-training mixed-precision quantization. Our approach is to train the random noise to generate data by using class-relevant vectors. It is not only independent of network structure but also provides a labeled dataset. We demonstrate that the generated data are sufficient to ensure the quality of post-training quantization. Furthermore, we proposed a novel sensitivity metric, which is important to optimize bit allocation. The proposed sensitivity metric considers the effect of quantization on other relative layers and task loss together using the gradient perturbation of the quantized neural network. Comparisons of sensitivity metrics were made to show the extent to which a layer with high sensitivity, measured with sensitivity metrics of other methods, affects task loss. The proposed sensitivity metric outperformed other metrics to stand for the effect of quantization error. We leave optimizing bit allocation using the proposed metric and applying to other tasks as part of future work.

\vfill\pagebreak

\bibliographystyle{ieeetran}
\bibliography{icip2021}

\end{document}